\RequirePackage{fix-cm}
\documentclass[sn-mathphys]{sn-jnl}

\usepackage{tabularx}
\usepackage{multirow}
\usepackage{setspace}
\usepackage{amsmath}
\usepackage{natbib}
\usepackage{subcaption}
\usepackage{graphicx}
\usepackage{adjustbox}
\usepackage{longtable}
\usepackage{amssymb}
\usepackage{array}
\usepackage{comment}
\usepackage{anyfontsize}
\usepackage{silence}
\usepackage{float} 

\jyear{2021}

\theoremstyle{thmstyleone}

\theoremstyle{thmstyletwo}

\theoremstyle{thmstylethree}

\begin{document}

\title[HieroGlyphTranslator: Automatic Translation of Hieroglyphs to English ] {HieroGlyphTranslator: Automatic Recognition and Translation of Egyptian Hieroglyphs to English}

\author[1]{\fnm{Ahmed} \sur{Nasser}}
\author[1]{\fnm{Marwan} \sur{Mohamed}}
\author[1]{\fnm{Alaa} \sur{Sherif}}
\author[1]{\fnm{Basmala} \sur{Mahmoud}}
\author[1]{\fnm{Shereen} \sur{Yehia}}
\author[1]{\fnm{Asmaa} \sur{Saad}}
\author[1]{\fnm{Mariam S.} \sur{El-Rahmany}}
\author*[1,2]{\fnm{Ensaf H.} \sur{Mohamed}}\email{enmohamed@nu.edu.eg}

\affil[1]{\orgdiv{Computer Science Department}, \orgname{Faculty of Computers and Artificial Intelligence}, \orgaddress{\street{Helwan University}, \city{Cairo}, \state{Egypt}}}
\affil[2]{\orgdiv{School of Information Technology and Computer Science (ITCS)}, \orgname{Nile University},\city{Giza},\state{Egypt}}


\abstract{Egyptian hieroglyphs, the ancient Egyptian writing system, are composed entirely of drawings. Translating these glyphs into English poses various challenges, including the fact that a single glyph can have multiple meanings. Deep learning translation applications are evolving rapidly, producing remarkable results that significantly impact our lives. In this research, we propose a method for the automatic recognition and translation of ancient Egyptian hieroglyphs from images to English. This study utilized two datasets for classification and translation: the Morris Franken dataset and the EgyptianTranslation dataset. Our approach is divided into three stages: segmentation (using Contour and Detectron2), mapping symbols to Gardiner codes, and translation (using the CNN model). The model achieved a BLEU score of 42.2, a significant result compared to previous research.}


\maketitle

\section{Introduction}

Nowadays, advancements in Artificial Intelligence (AI), particularly in machine and deep learning, give new potential to create tools that help the work of specialists in fields that appear to be unrelated to information technology. Ancient Egyptian hieroglyphic writing is one example of such a field.

The automatic recognition and translation of Egyptian hieroglyphs is an innovative and interesting topic of study that integrates computer science, linguistics, and Egyptology. Egyptian hieroglyphs, which date back to 3000 BCE, are one of the world's earliest writing systems. 

\begin{figure}[h]
    \centering
    \includegraphics[width=\linewidth]{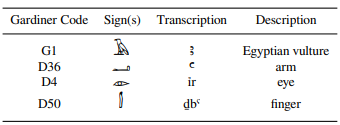}
    \caption{Example of Egyptian literals \cite{Wiesenbach2019}}
    \label{fig:example_literals}
\end{figure}

[Image of Ancient Egyptian Hieroglyphs on a wall]
 However, due to the complexity of the hieroglyphic character and a lack of experts in the field, translating ancient texts can be difficult and time-consuming. The creation of an automatic recognition and translation system for Egyptian hieroglyphs has the potential to transform Egyptology and make these ancient writings more accessible to a wider audience. A hieroglyph can represent any of the following (Figure~\ref{fig:example_literals}).

The Egyptian hieroglyph is a complicated symbol made up of two parts: an ideogram and a phonogram. An ideogram (Semagram) is a graphic symbol that represents a notion in relation to another. The phonogram can also play two roles: the proper phonogram, which can indicate the phonetic value of the sign as well as a phonetic sequence, and the phonetic complement, which is a specific series of signs that expresses the sound of the sign to which it relates in a redundant way (see Fig.~\ref{fig:hieroglyph_structure}). Given its complexity, the hieroglyphic symbol provides a rich ground for the application of a deep learning approach to its recognition and classification \cite{Barucci2021}.

\begin{figure}[h]
    \centering
    \includegraphics[scale=0.7]{hieroglyph_structure.png}
    \caption{The structure of an Egyptian hieroglyph: (a) Semagram (Owl) and phonogram /m/; (b) Semagram (Nose), determinative (nose, face, and associated actions), and phonogram; (c) Semagram (Mouth), phonogram /r/ and phonetic complement (R).}
    \label{fig:hieroglyph_structure}
\end{figure}

The proposed system in this research is an automatic recognition and translation system for Egyptian hieroglyphs named "The Rosetta Translation Application", aimed to help scholars, researchers, and visitors interpret and evaluate ancient manuscripts. To recognize and translate hieroglyphs into English, the application employs deep learning algorithms and a vast training sample of hieroglyphic texts and images. The system is available via a user-friendly interface, allowing users to capture images of hieroglyphic text and obtain an exact English translation in real-time.

The main contributions of this paper are summarized as follows:
\begin{itemize}
    \item Develop an automatic recognition and translation system for Egyptian hieroglyphs using a CNN model and computer vision techniques.
    \item Create and collect a large dataset of hieroglyphic texts and images to be used for training and evaluating the system.
    \item Evaluate the accuracy and effectiveness of the system in recognizing and translating hieroglyphic text using established benchmarks and metrics for performance.
    \item Address the technical challenges, such as handling variations in spelling and grammar, dealing with many symbols and their variations, and processing images of hieroglyphic inscriptions.
\end{itemize}

This paper is structured as follows: Section 2 discusses related work. Section 3 describes our proposed model, techniques, datasets, and models. Section 4 presents the experimental work and results. Finally, Section 5 offers a conclusion and future work.

\section{Related Work}

In this section, we investigate the related work in the field of automatic recognition and translation of Egyptian hieroglyphs.

Andrea Barucci et al. \cite{Barucci2021} utilized various Convolutional Neural Networks (CNNs) to address the problem of ancient Egyptian hieroglyphs classification. They faced challenges in classifying images of hieroglyphs from two different datasets. Three well-known CNN architectures (ResNet-50, Inception-v3, and Xception) were tested, in addition to a dedicated CNN named Glyphnet. Glyphnet demonstrated the best performance with 97.6\% accuracy.

Franken and Van Gemert \cite{Franken2013} presented an automated transliteration system utilizing five popular image descriptors (HOG, HOOSC, HOOSS, ShapeContext, and Self Similarity) and three common matching schemes. They employed two language models (Lexicon and N-gram) to consider the context. The model achieved an average score of 74\% with manually annotated hieroglyphs and 53\% with automatically annotated hieroglyphs.

Reham Elnabawy et al. \cite{Elnabawy2018} utilized the Optical Character Recognition (OCR) algorithm. They applied image segmentation and the HOG matching technique, linking Gardiner's code to improve image recognition. Their method achieved an accuracy of 66.64\%.

Tommaso Guidi \cite{Guidi2023} utilized Detectron2, which includes different R-CNN models for image segmentation. Their system is composed of two modules: the Region Proposal Network (RPN) and a second module that classifies the proposed regions. ResNet, pre-trained on the COCO dataset, was employed as the backbone network.

Ragaa Moustafa et al. \cite{Moustafa2022} proposed a Flutter-based mobile application called Scriba that uses deep learning techniques to identify and translate hieroglyphic text. They tested different deep learning models and found that MobileNet and EfficientNet achieved higher accuracy.

Asmaa Sobhy et al. \cite{Sobhy2023} proposed a deep learning model applying image processing, NLP techniques, and the Siamese network. Their model detects glyphs with a mean Average Precision (mAP) of 95\% and an Average Recall (AR) of 75\%.

\begin{table}[htbp]
    \centering
    \caption{Comparison between related work}
    \label{table:comparison}
    \begin{tabularx}{\textwidth}{|p{2cm}|X|X|}
        \hline
        \textbf{Ref.} & \textbf{Strengths} & \textbf{Weakness} \\
        \hline
        \cite{Barucci2021} & High accuracy (97.6\%) with multiple CNN architectures (Inception-v3, Xception, ResNet50) and Glyphnet model. & Focused on single hieroglyphs. \\
        \hline
        \cite{Franken2013} & Used HOG descriptor, spatial matching, and N-gram for context-aware recognition; achieved 53\% average score. & Manual descriptors outperformed automatic approach. \\
        \hline
        \cite{Elnabawy2018} & Enhanced detection accuracy with HOG matching and Gardiners code; achieved 66.64\%. & OCR limitations, e.g., 'rn' misread as 'm'; poor English sentence construction. \\
        \hline
        \cite{Guidi2023} & High object detection accuracy (98.4\%) using Mask R-CNN with bounding box and mask regression. & Small sample size; poor segmentation of low contrast glyphs. \\
        \hline
        \cite{Moustafa2022} & Improved CNN accuracy with data augmentation. & Limited dataset (316 images); some misclassifications. \\
        \hline
        \cite{Sobhy2023} & High mAP (95\%) and AR (75\%) with Siamese network; BLEU score of 59.19 for translation. & Missing characters require removal or expert interpretation; non-modern English corpus. \\
        \hline
        \cite{Tang2016} & High accuracy (88.56\%) using CNN-based transfer learning. & Small dataset; misrecognition of similar forms and low-quality sources. \\
        \hline
        \cite{AlNoori2020} & 100\% accuracy in Sumerian character recognition using CNN and region splitting technique. & Small dataset. \\
        \hline
    \end{tabularx}
\end{table}

Tang et al. \cite{Tang2016} proposed a CNN-based transfer learning method for recognizing historical Chinese characters. Al-Noori et al. \cite{AlNoori2020} proposed a model for extracting Sumerian characters using CNN and Region Splitting, achieving 100\% accuracy in recognition for each extracted character. Asmaa Sobhy et al. \cite{AsmaaSobhy2023} used a hierarchical classification approach with ResNet50 and a Siamese network, achieving a glyph detection accuracy of 95\% and a translation BLEU score of 59.19.

\section{Materials and Methods}

In the following subsections, the datasets and proposed models are explained in detail.

\subsection{Datasets}

\begin{figure}[htbp]
        \centering
        \includegraphics[width=0.45\textwidth]{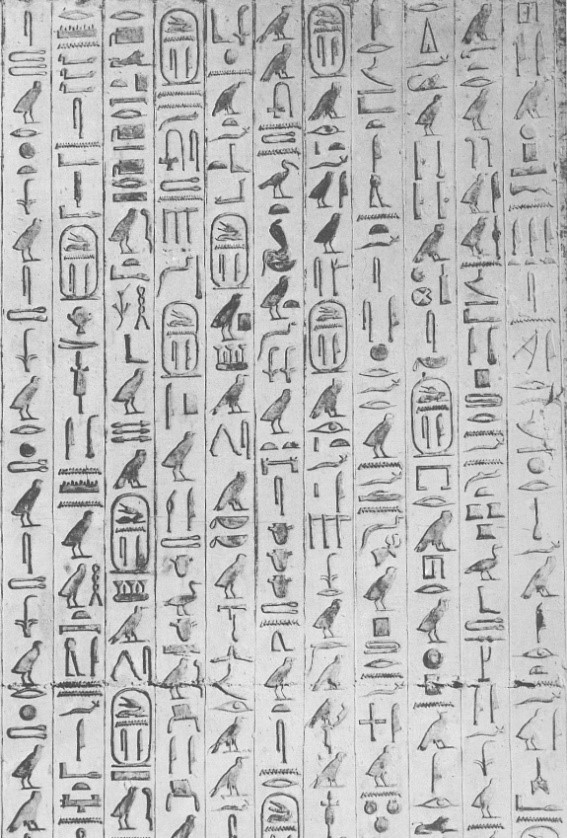}
        \caption{Example piece of Unas pyramids wall}
        \label{fig:UnasSample}
    \end{figure}

\begin{figure}[htbp]
    \centering
    \begin{subfigure}[b]{0.7\textwidth}
        \centering
        \includegraphics[width=\textwidth]{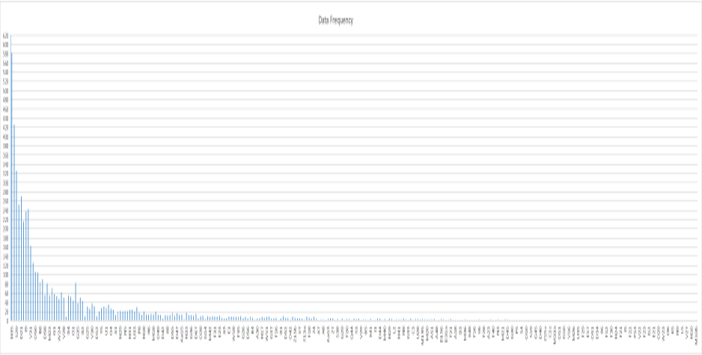}
        \caption{Dataset distribution before image augmentation}
        \label{fig:DatasetDistBefore}
    \end{subfigure}
    \vfill
    \begin{subfigure}[b]{0.7\textwidth}
        \centering
        \includegraphics[width=\textwidth]{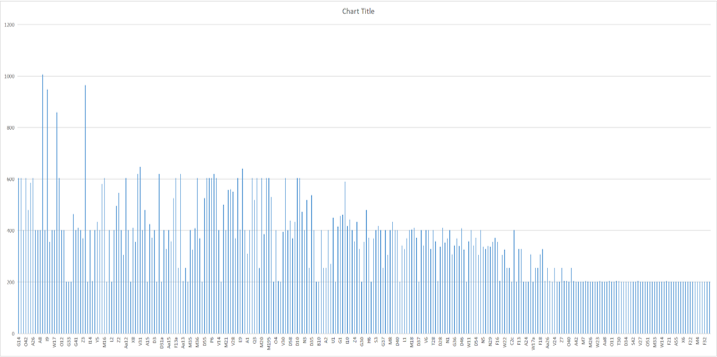}
        \caption{Dataset distribution after image augmentation}
        \label{fig:DatasetDistAfter}
    \end{subfigure}
    \caption{Data distribution before and after augmentation}
    \label{fig:DatasetDist}
\end{figure}

In this research, we utilized three datasets as follows:

\begin{enumerate}
    \item 51 images from \cite{AdobeStock}, \cite{AlamyStock} for segmentation (specifically for Detectron2 \& Contour). Using the LabelMe tool, we annotated the photographs and saved the files in JSON format.

    \item The Morris Franken dataset, commonly known as the Hieroglyphic Dataset Pyramid of Unas \cite{Piankoff1969}. Both a visual set and a textual corpus are included. Ten plate images of hieroglyphs, totaling 4210 glyphs and 171 distinct hieroglyphs, make up the visual collection. Figure \ref{fig:UnasSample} depicts a sample segment of the Unas pyramid. In this collection, each glyph is tagged with its Gardiner's code . A selection of the dataset's 171 distinct Gardiner's codes is displayed in Table \ref{table:gardinerCodes}.

    \item In the Pyramid of Unas dataset, there are more than 700 distinct hieroglyphic characters. We introduced 120 new classes with 1220 newly collected photographs. As a result, our final dataset, which is depicted in Figure \ref{fig:DatasetDist}, has 5430 samples spread among 291 classes. By applying data augmentation techniques, the dataset expanded to 102,401 images.

    \item The Ancient Egyptian texts and their translations are included in the EgyptianTranslation dataset \cite{RoseGitHub}. There are 150 texts in all (12,938 sentences) in the collection, divided into Funerary, Literary, and Historical texts.
\end{enumerate}

\begin{table}[h!]
\centering
\caption{Sample of the Dataset}
\begin{tabular}{| >{\centering\arraybackslash}m{1.5cm} | >{\centering\arraybackslash}m{1.5cm} | >{\centering\arraybackslash}m{1.5cm} | >{\centering\arraybackslash}m{1.5cm} | >{\centering\arraybackslash}m{1.5cm} |}
\hline
\textbf{Gardiner Code} & V31 & Z1 & M17 & I9 \\
\hline
\textbf{Image} & \includegraphics[width=1cm]{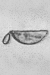} & \includegraphics[width=1cm]{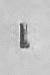} & \includegraphics[width=1cm]{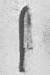} & \includegraphics[width=1cm]{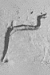} \\
\hline
\end{tabular}
\label{table:gardinerCodes}
\end{table}

\subsection{The Proposed Pipeline Architecture}

\begin{figure}[htbp]
    \centering
    \includegraphics[width=0.9\textwidth]{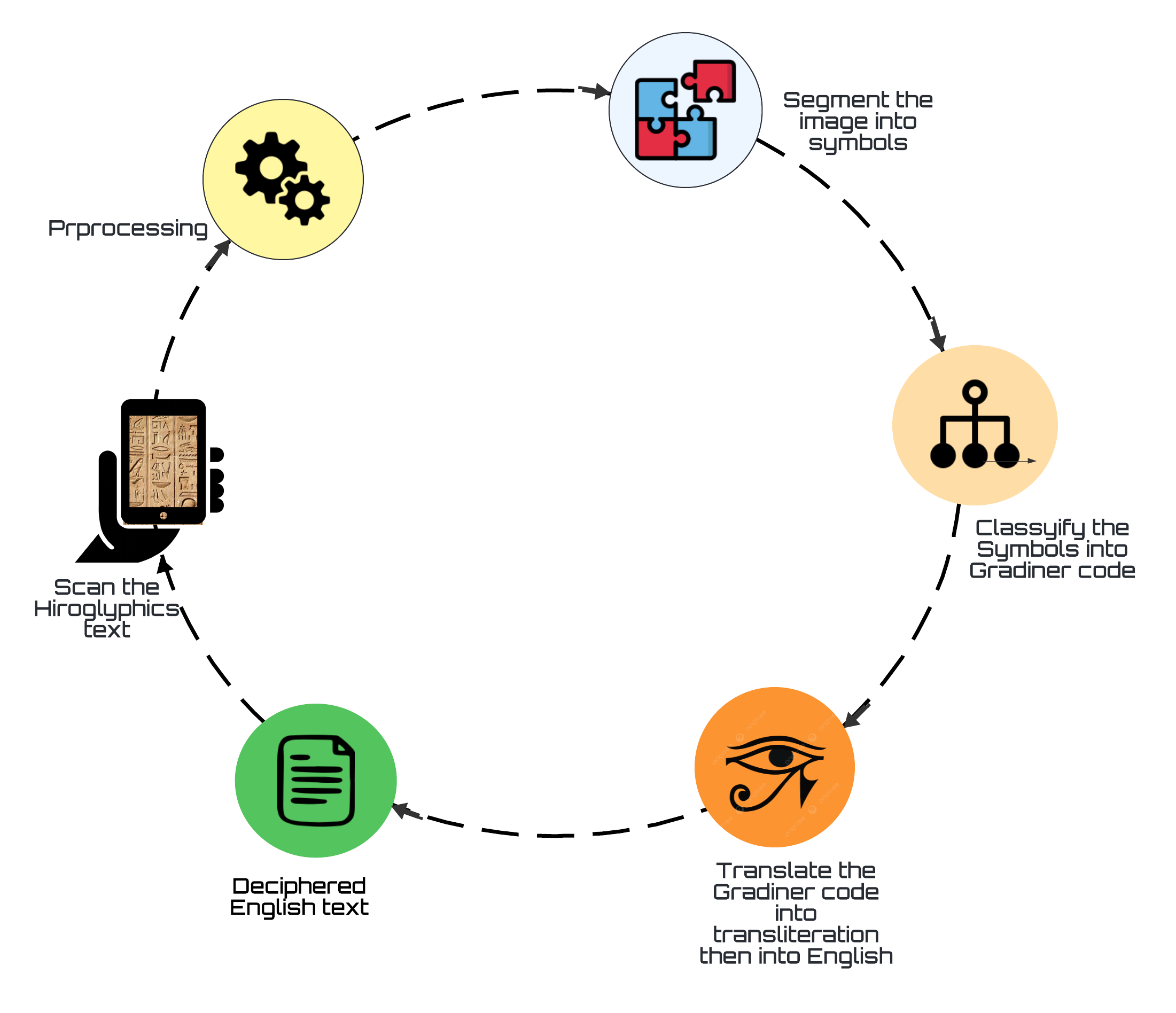}
    \caption{The proposed pipeline architecture}
    \label{fig:pipeline_architecture}
\end{figure}

The proposed pipeline model consists of four main models: Segmentation, Classification, and Translation, in addition to Preprocessing as shown in main pipeline Figure \ref{fig:pipeline_architecture}.

\subsubsection{Preprocessing}

In this study, we implemented several image processing techniques. The Hough transform \cite{Hough2015} was utilized to separate the columns in the input images .

\begin{figure}[htbp]
    \centering
    \begin{subfigure}[b]{0.45\textwidth}
        \centering
        \includegraphics[width=\textwidth]{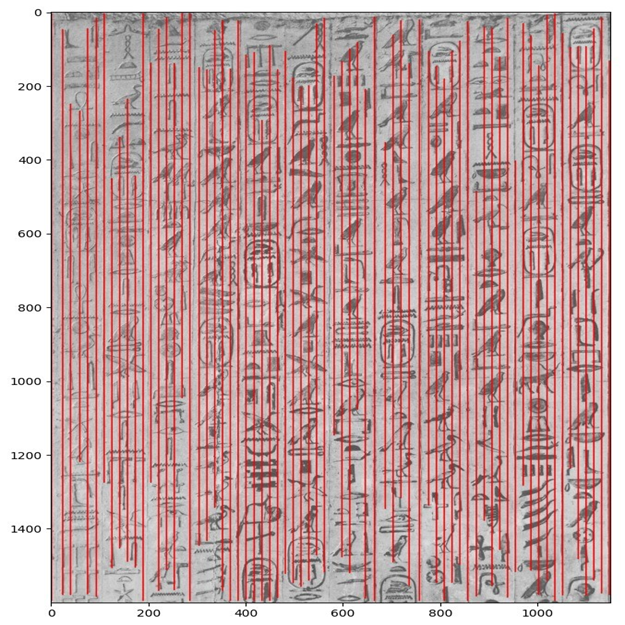}
        \caption{1: Detection of multiple lines}
        \label{fig:hough_problem1}
    \end{subfigure}
    \hfill
    \begin{subfigure}[b]{0.45\textwidth}
        \centering
        \includegraphics[width=\textwidth]{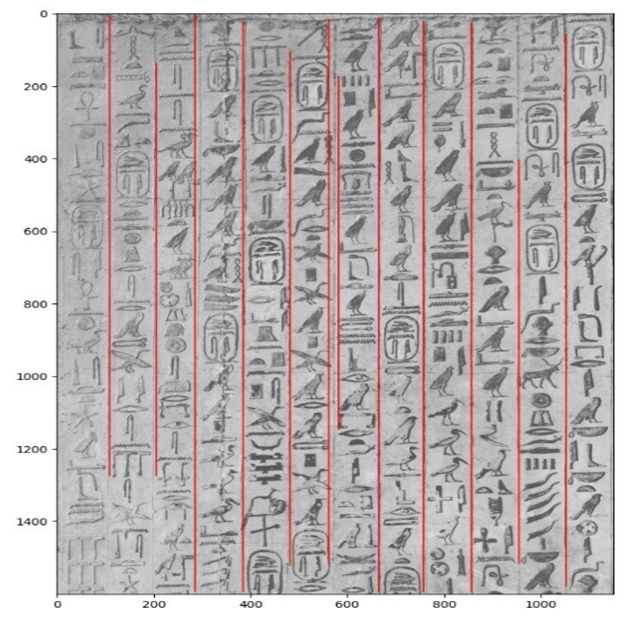}
        \caption{2: Detection of incomplete lines}
        \label{fig:hough_problem2}
    \end{subfigure}
    \begin{subfigure}[b]{0.45\textwidth}
        \centering
        \includegraphics[width=\textwidth]{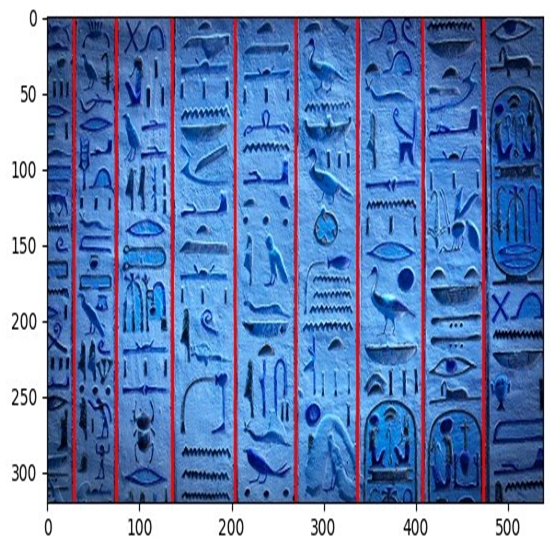}
        \caption{3: Unequal columns or rows}
        \label{fig:unequal_columns}
    \end{subfigure}
    \caption{Problems encountered with the Hough transform technique}
    \label{fig:hough_problems}
\end{figure}

\textbf{Problem 1: Detection of Irrelevant Lines}
The primary issue was the detection of multiple lines within an image, including irrelevant lines derived from hieroglyphic symbols as shown in Figure \ref{fig:hough_problem1}.

\textbf{Solution to Problem 1: Filtering Relevant Lines}
We developed a function to divide the image's width by the number of rows/columns. Each line detected by the Hough transform was then compared to the lines derived from this function.

\textbf{Problem 2: Detection of Incomplete Lines}
Another problem was the detection of incomplete lines that did not reach to the edges of the image as shown in Figure \ref{fig:hough_problem2}.

\textbf{Solution to Problem 2: Post-Processing Refinement}
We initialized variables for cropping limits based on the image size to ensure the Hough transform only identified lines within boundaries.

\textbf{Problem 3: Unequal Columns or Rows}
We also encountered the issue of unequal columns (or rows), as shown in Fig. \ref{fig:unequal_columns}.

\textbf{Solution to Problem 3: Equalizing Columns and Rows}
By developing a function that divides the image's width/height by the number of columns/rows, we ensured consistent separation.

 \begin{figure}[htbp]
        \centering
        \includegraphics[width=0.5\textwidth]{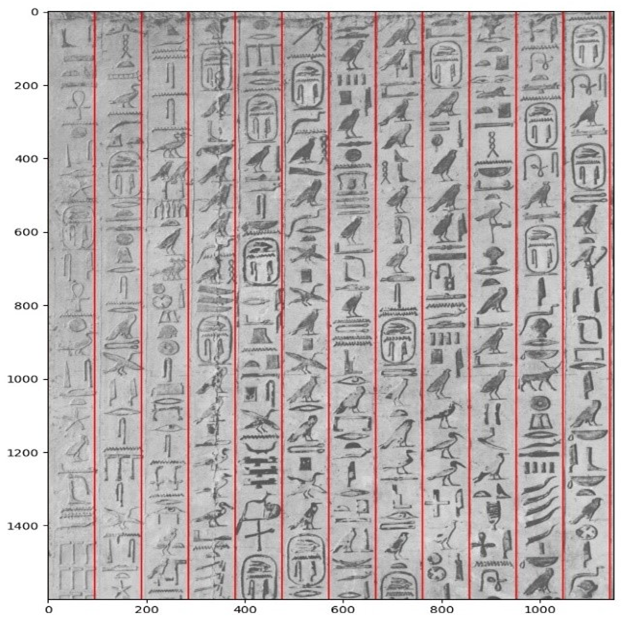}
        \caption{An image after preprocessing}
        \label{fig:preprocessing_result}
    \end{figure}

\subsubsection{Segmentation Module}

We deployed a hybrid method combining contour detection and Detectron2.

\begin{figure}[htbp]
    \centering
    \begin{subfigure}[b]{0.45\textwidth}
        \centering
        \includegraphics[width=\textwidth]{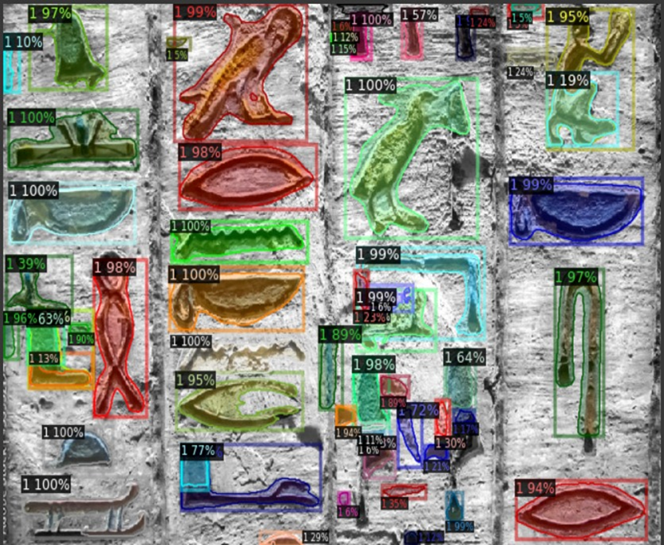}
        \caption{Result of applying Detectron2}
        \label{fig:detectron2_sample}
    \end{subfigure}
    \hfill
    \begin{subfigure}[b]{0.45\textwidth}
        \centering
        \includegraphics[width=\textwidth]{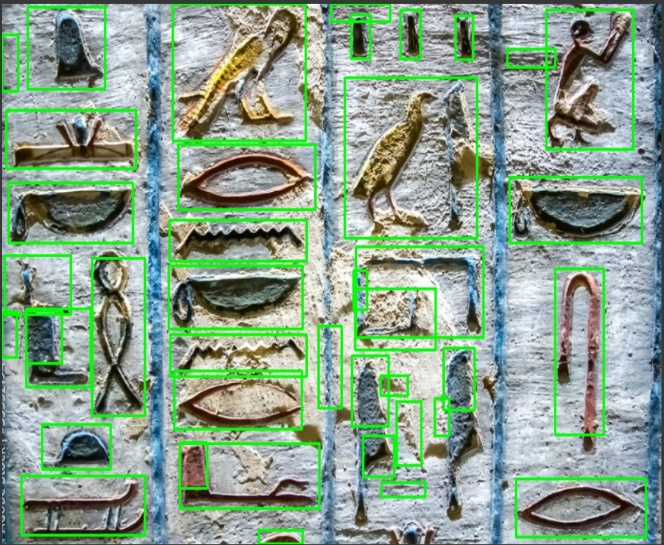}
        \caption{Hybrid of Detectron2 \& Contour}
        \label{fig:hybrid_sample}
    \end{subfigure}
\end{figure}

\begin{itemize}
    \item \textbf{Contour Detection:} \cite{Gong2018} This technique is used to find boundaries or edges in an image by producing a binary image.
    \item \textbf{Detectron2:} \cite{Guidi2023} Detectron2 is a state-of-the-art framework for object recognition and image segmentation. We used the Roboflow tool to annotate our segmentation dataset.
    \item \textbf{Hybrid Contour and Detectron2:} To enhance our work, we combined Contour detection and Detectron2, as shown in Fig. \ref{fig:hybrid_sample}.
\end{itemize}

\subsubsection{Classification Module}

We used ResNet50 \cite{he2016deep} 

\begin{figure}[htbp]
    \centering
    \begin{subfigure}[b]{\textwidth}
        \centering
        \includegraphics[width= \textwidth]{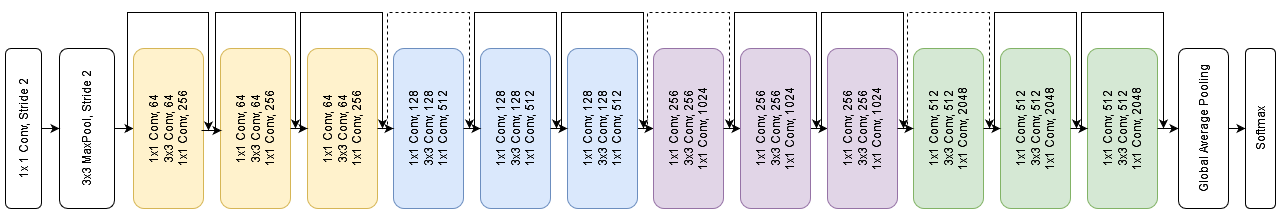}
        \caption{ResNet50 architecture \cite{he2016deep}}
        \label{fig:resnet50}
    \end{subfigure}
    \hfill
    \begin{subfigure}[b]{\textwidth}
        \centering
        \includegraphics[width=.25 \textwidth]{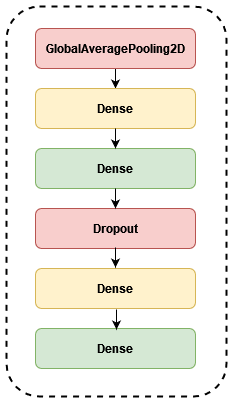}
        \caption{Modified ResNet50 architecture}
        \label{fig:resnet50_adaptation}
    \end{subfigure}
    \caption{The original and adapted ResNet50 architectures.}
    \label{fig:resnet50_figures}
\end{figure}

[Image of ResNet50 architecture diagram]
, a deep neural network architecture, as the pre-trained model for image classification tasks. To adapt ResNet50 for our task, we employed transfer learning by adding layers to the base model as shown in Figure \ref{fig:resnet50_figures}.

\subsubsection{Transliteration and Translation Module}

This module consisted of two processes: converting Gardiner code to transliteration, and translating transliteration to English using a Sequence-to-Sequence model 

\begin{figure}[htbp]
    \centering
    \includegraphics[width=\textwidth]{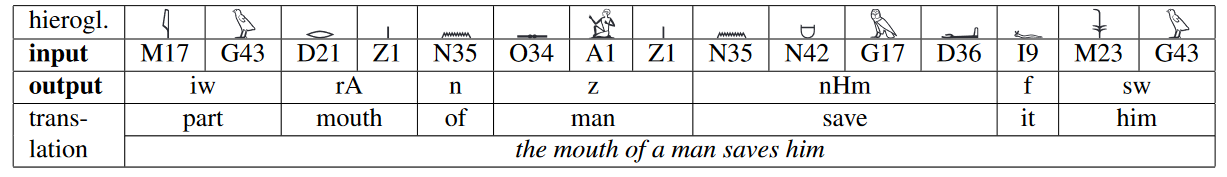}
    \caption{Example of the image translation process: converting Gardiner code to transliteration, and transliteration to English \cite{FSMNLP2011}}
    \label{fig:translation_example}
\end{figure}

[Image of Transformer model architecture]
 (Fig. \ref{fig:translation_example}). We employed the transformer architecture based on the OpenNMT framework.

\begin{figure}[htbp]
    \centering
    \includegraphics[width=0.8\textwidth]{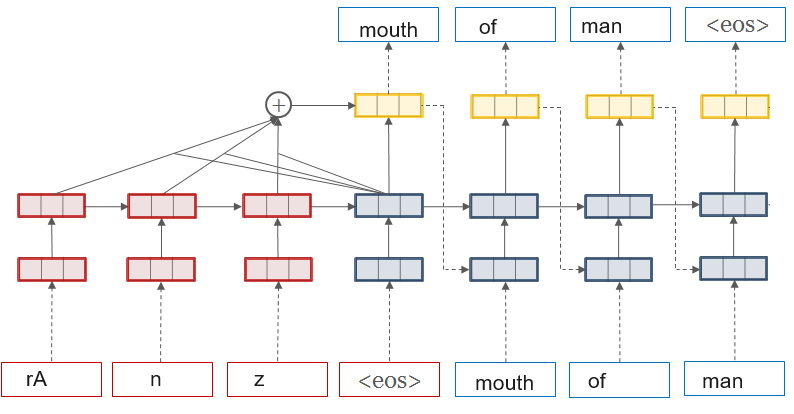}
    \caption{The translation model based on Encoder-Decoder architecture.}
    \label{fig:translation_model}
\end{figure}

\section{Results and Discussion}\label{section4-2}

\subsection{Evaluation Metrics}

The performance is evaluated using Accuracy, Precision, Recall, F1 Score, Perplexity (PPL), and BLEU score.

\begin{equation}
\text{Accuracy} = \frac{TP + TN}{TP + TN + FP + FN}
\label{eq:accuracy}
\end{equation}

\begin{equation}
\text{Precision} = \frac{TP}{TP + FP}
\label{eq:precision}
\end{equation}

\begin{equation}
\text{Recall} = \frac{TP}{TP + FN}
\label{eq:recall}
\end{equation}

\begin{equation}
\text{F1 Score} = 2 \cdot \frac{\text{Precision} \cdot \text{Recall}}{\text{Precision} + \text{Recall}}
\label{eq:f1}
\end{equation}

\subsection{Experimental Results}

\subsubsection{Classification Module}

\begin{figure}[htbp]
    \centering
    \begin{subfigure}[b]{0.45\textwidth}
        \centering
        \includegraphics[width=\textwidth]{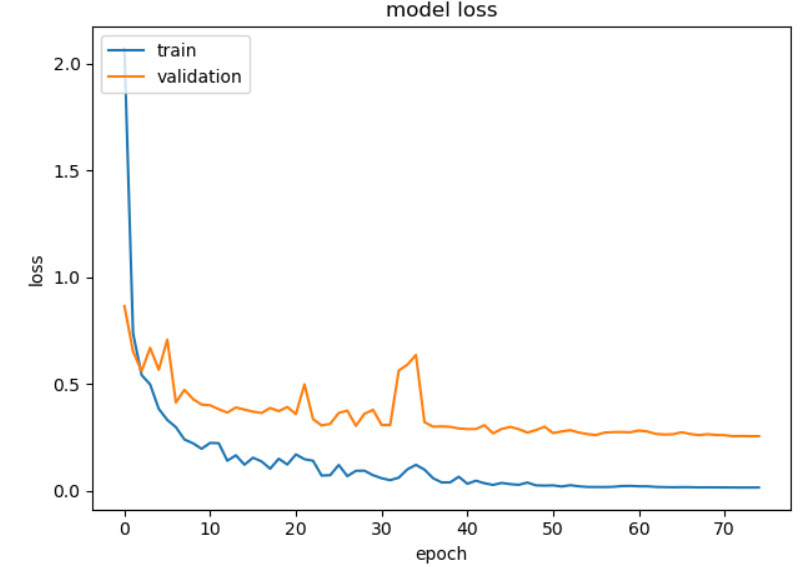}
        \caption{Classification Loss}
        \label{fig:classification_loss}
    \end{subfigure}
    \hfill
    \begin{subfigure}[b]{0.45\textwidth}
        \centering
        \includegraphics[width=\textwidth]{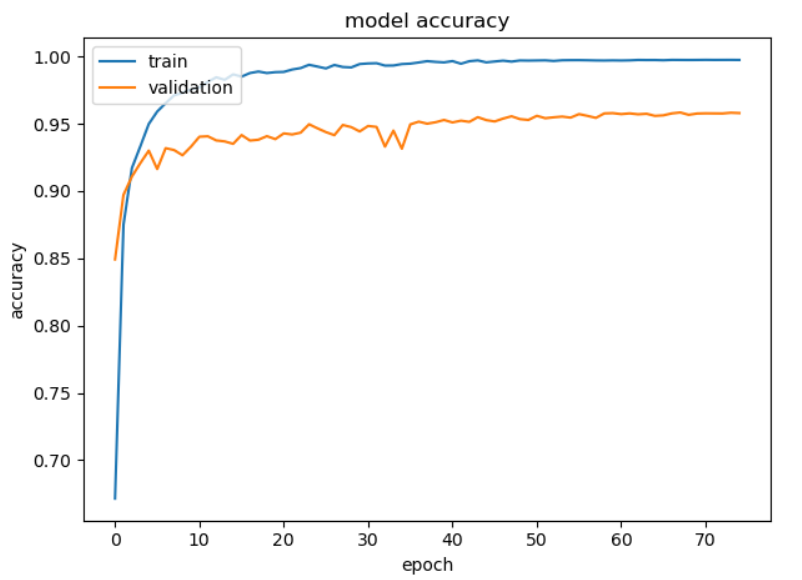}
        \caption{Classification Accuracy}
        \label{fig:classification_accuracy}
    \end{subfigure}
    \caption{Results of the Classification Model}
    \label{fig:classification_results}
\end{figure}

We separated the dataset into 60\% training, 20\% validation, and 20\% testing. Results are depicted in Fig. \ref{fig:classification_results}.

\begin{table}[ht]
    \centering
    \caption{Classification Model Results - Accuracy and Loss}
    \label{tab:classification_results_1}
    \begin{tabularx}{\textwidth}{|c|X|X|X|X|X|X|}
        \hline
        \textbf{Classes} & \textbf{Training Acc.} & \textbf{Training Loss} & 
        \textbf{Validation Acc.} & \textbf{Validation Loss} & 
        \textbf{Test Acc.} & \textbf{Test Loss} \\
        \hline
        291 & 99.74\% & 0.16 & 95.79\% & 0.2564 & 81.8\% & 0.23 \\
        83  & 97\%    & 0.17 & 92\%    & 0.41   & 91\%   & 0.3695 \\
        \hline
    \end{tabularx}
\end{table}

\begin{table}[ht]
    \centering
    \caption{Classification Model Results - Precision and Recall}
    \label{tab:classification_results_2}
    \begin{tabularx}{0.8\textwidth}{|c|X|X|}
        \hline
        \textbf{Classes} & \textbf{Precision} & \textbf{Recall} \\
        \hline
        291 & 0.8724 & 0.8052 \\
        83  & 0.9396 & 0.9125 \\
        \hline
    \end{tabularx}
\end{table}

\subsubsection{Translation Module}

The "PRED AVG SCORE" obtained was 0.3651, as shown in Table \ref{tab:pred_avg_score_ppl_results}. We found that our system achieved a better BLEU Score (42.22) compared to the baseline system \cite{Wiesenbach2019} as shown in Table \ref{tab:BLEU_score_results}.

\begin{table}[ht]
    \centering
    \caption{PRED AVG SCORE and PRED PPL Results}
    \label{tab:pred_avg_score_ppl_results}
    \begin{tabularx}{0.8\textwidth}{|X|X|}
        \hline
        \textbf{PRED AVG SCORE} & \textbf{PRED PPL} \\
        \hline
        0.3651 & 1.4407 \\
        \hline
    \end{tabularx}
\end{table}

\begin{table}[ht]
    \centering
    \caption{BLEU Score Results}
    \label{tab:BLEU_score_results}
    \begin{tabularx}{0.8\textwidth}{|X|X|X|X|X|}
        \hline
        \textbf{Model} & \textbf{BLEU SCORE} & \textbf{NLTK's BLEU} & \textbf{NLTK's corpus BLEU} & \textbf{sacreBLEU-BLEU} \\
        \hline
        Wiesenbach, P. \& Riezler, S.~\citep{Wiesenbach2019} & 22.38 & - & - & - \\
        \hline
        Proposed method & 42.22 & 21.29 & 35.68 & 63.89 \\
        \hline
    \end{tabularx}
\end{table}

The study \cite{AsmaaSobhy2023} notes that the lack of language resources results in distorted English phrases. Our evaluation of random samples is shown in Table \ref{tab:loss_acc_test_results} and Table \ref{tab:comparison_translations}.

\begin{table}[ht]
    \centering
    \caption{Loss and Accuracy of Test Data for Translation Module}
    \label{tab:loss_acc_test_results}
    \begin{tabularx}{0.8\textwidth}{|X|X|}
        \hline
        \textbf{Metric} & \textbf{Value} \\
        \hline
        Loss & 0.3695 \\
        \hline
        Accuracy & 0.9177 \\
        \hline
    \end{tabularx}
\end{table}

\begin{table}[ht]
\centering
\caption{Comparison of Target and Predicted Translations}
\label{tab:comparison_translations}
\begin{tabularx}{\textwidth}{|X|X|}
\hline
\textbf{The Target Translations} & \textbf{The Predicted Translations} \\
\hline

\begin{tabular}[t]{@{}l@{}}
-- are you not a man \\
-- has the house been prepared \\
-- the keeper of the diadem \\
-- reception of the tribute of a prince of punt \\
-- makes everyone laugh by his evil wrongdoing \\
-- like the sky's clearing \\
-- and they put the barley in a sealed room \\
-- and justice was established in its place \\
-- who enters to report \\
-- so that his hens came to rest on the water \\
-- and was very frightened
\end{tabular}
&
\begin{tabular}[t]{@{}l@{}}
-- are you not a man \\
-- is it the house has gone out \\
-- the keeper of the diadem \\
-- Ssp of the tribute of a prince of punt \\
-- makes everyone ,f by his evil wrongdoing \\
-- like the sky's clearing \\
-- and they put the barley here in a room \\
-- while justice has pain in its place \\
-- upon reporting the collection \\
-- so that its xnn came to rest on the water \\
-- and was very frightened
\end{tabular}
\\
\hline

\begin{tabular}[t]{@{}l@{}}
-- as you wish that your city gods praise you \\
-- and forget worries \\
-- with a pedestal of silver \\
-- it is that there is fright in my body \\
-- who passes eternity \\
-- and the loved man for whom a city is founded \\
-- i didn't know whether i was dead or alive \\
-- as commander of his army \\
-- then it is ground \\
-- under the majesty of
\end{tabular}
&
\begin{tabular}[t]{@{}l@{}}
-- as you praise and as you love your city gods \\
-- and forget worries \\
-- and a pedestal of silver \\
-- this is what keeps her in my belly \\
-- who passes eternity \\
-- founded for the city of n \\
-- that i could distinguish life from me \\
-- on the commander of his army \\
-- then it is ground \\
-- under the majesty of
\end{tabular}
\\
\hline
\end{tabularx}
\end{table}

\section{Conclusion}

The proposed model represents a significant advancement in the field of Automatic Recognition and Translation of Egyptian Hieroglyphs to English. By integrating cutting-edge machine learning and computer vision techniques with a deep understanding of ancient Egyptian culture and history, this app provides a powerful and accessible tool for exploring hieroglyphics.

\backmatter

\section*{Declarations}
\section*{Ethics approval and consent to participate}
\section*{Consent for publication}
Not applicable.

\section*{Competing interests}
The authors declare that they have no competing interests.

\section*{Authors' contributions}
All authors listed have made a substantial, direct, and intellectual contribution to the work and approved it for publication.

\section*{Funding}
Not applicable.

\section*{Acknowledgements}
Not applicable.

\section*{Availability of data and materials}
The data referenced in this article is openly and freely accessible via this link \cite{OurDataset}

\bibliography{ref}


\begin{thebibliography}{19}
\ifx \bisbn   \undefined \def \bisbn  #1{ISBN #1}\fi
\ifx \binits  \undefined \def \binits#1{#1}\fi
\ifx \bauthor  \undefined \def \bauthor#1{#1}\fi
\ifx \batitle  \undefined \def \batitle#1{#1}\fi
\ifx \bjtitle  \undefined \def \bjtitle#1{#1}\fi
\ifx \bvolume  \undefined \def \bvolume#1{\textbf{#1}}\fi
\ifx \byear  \undefined \def \byear#1{#1}\fi
\ifx \bissue  \undefined \def \bissue#1{#1}\fi
\ifx \bfpage  \undefined \def \bfpage#1{#1}\fi
\ifx \blpage  \undefined \def \blpage #1{#1}\fi
\ifx \burl  \undefined \def \burl#1{\textsf{#1}}\fi
\ifx \doiurl  \undefined \def \doiurl#1{\url{https://doi.org/#1}}\fi
\ifx \betal  \undefined \def \betal{\textit{et al.}}\fi
\ifx \binstitute  \undefined \def \binstitute#1{#1}\fi
\ifx \binstitutionaled  \undefined \def \binstitutionaled#1{#1}\fi
\ifx \bctitle  \undefined \def \bctitle#1{#1}\fi
\ifx \beditor  \undefined \def \beditor#1{#1}\fi
\ifx \bpublisher  \undefined \def \bpublisher#1{#1}\fi
\ifx \bbtitle  \undefined \def \bbtitle#1{#1}\fi
\ifx \bedition  \undefined \def \bedition#1{#1}\fi
\ifx \bseriesno  \undefined \def \bseriesno#1{#1}\fi
\ifx \blocation  \undefined \def \blocation#1{#1}\fi
\ifx \bsertitle  \undefined \def \bsertitle#1{#1}\fi
\ifx \bsnm \undefined \def \bsnm#1{#1}\fi
\ifx \bsuffix \undefined \def \bsuffix#1{#1}\fi
\ifx \bparticle \undefined \def \bparticle#1{#1}\fi
\ifx \barticle \undefined \def \barticle#1{#1}\fi
\bibcommenthead
\ifx \bconfdate \undefined \def \bconfdate #1{#1}\fi
\ifx \botherref \undefined \def \botherref #1{#1}\fi
\ifx \url \undefined \def \url#1{\textsf{#1}}\fi
\ifx \bchapter \undefined \def \bchapter#1{#1}\fi
\ifx \bbook \undefined \def \bbook#1{#1}\fi
\ifx \bcomment \undefined \def \bcomment#1{#1}\fi
\ifx \oauthor \undefined \def \oauthor#1{#1}\fi
\ifx \citeauthoryear \undefined \def \citeauthoryear#1{#1}\fi
\ifx \endbibitem  \undefined \def \endbibitem {}\fi
\ifx \bconflocation  \undefined \def \bconflocation#1{#1}\fi
\ifx \arxivurl  \undefined \def \arxivurl#1{\textsf{#1}}\fi
\csname PreBibitemsHook\endcsname

\bibitem{Wiesenbach2019}
\begin{bchapter}
\bauthor{\bsnm{Wiesenbach}, \binits{P.}},
\bauthor{\bsnm{Riezler}, \binits{S.}}:
\bctitle{Multi-task modeling of phonographic languages: Translating middle egyptian hieroglyphs}.
In: \bbtitle{Proceedings of the 16th International Conference on Spoken Language Translation}
(\byear{2019})
\end{bchapter}
\endbibitem

\bibitem{Barucci2021}
\begin{barticle}
\bauthor{\bsnm{Barucci}, \binits{A.}},
\bauthor{\bsnm{Cucci}, \binits{C.}},
\bauthor{\bsnm{Franci}, \binits{M.}},
\bauthor{\bsnm{Loschiavo}, \binits{M.}},
\bauthor{\bsnm{Argenti}, \binits{F.}}:
\batitle{A deep learning approach to ancient egyptian hieroglyphs classification}.
\bjtitle{IEEE Access}
\bvolume{9},
\bfpage{123438}--\blpage{123447}
(\byear{2021}).
\doiurl{10.1109/access.2021.3110082}
\end{barticle}
\endbibitem

\bibitem{Franken2013}
\begin{bchapter}
\bauthor{\bsnm{Franken}, \binits{M.}},
\bauthor{\bsnm{Van~Gemert}, \binits{J.C.}}:
\bctitle{Automatic egyptian hieroglyph recognition by retrieving images as texts}.
In: \bbtitle{Proceedings of the 21st ACM International Conference on Multimedia}
(\byear{2013}).
\doiurl{10.1145/2502081.2502199}
\end{bchapter}
\endbibitem

\bibitem{Elnabawy2018}
\begin{bchapter}
\bauthor{\bsnm{Elnabawy}, \binits{R.H.}},
\bauthor{\bsnm{Elias}, \binits{R.}},
\bauthor{\bsnm{Salem}, \binits{M.A.}}:
\bctitle{Image based hieroglyphic character recognition}.
In: \bbtitle{14th International Conference on Signal-Image Technology \& Internet-Based Systems (SITIS)}.
\bpublisher{IEEE}, \blocation{???}
(\byear{2018}).
\doiurl{10.1109/sitis.2018.00016}
\end{bchapter}
\endbibitem

\bibitem{Guidi2023}
\begin{barticle}
\bauthor{\bsnm{Guidi}, \binits{T.}},
\bauthor{\bsnm{Python}, \binits{L.}},
\bauthor{\bsnm{Forasassi}, \binits{M.}},
\bauthor{\bsnm{Cucci}, \binits{C.}},
\bauthor{\bsnm{Franci}, \binits{M.}},
\bauthor{\bsnm{Argenti}, \binits{F.}},
\bauthor{\bsnm{Barucci}, \binits{A.}}:
\batitle{Egyptian hieroglyphs segmentation with convolutional neural networks}.
\bjtitle{Algorithms}
\bvolume{16}(\bissue{2}),
\bfpage{79}
(\byear{2023}).
\doiurl{10.3390/a16020079}
\end{barticle}
\endbibitem

\bibitem{Moustafa2022}
\begin{bchapter}
\bauthor{\bsnm{Moustafa}, \binits{R.}},
\bauthor{\bsnm{Hesham}, \binits{F.}},
\bauthor{\bsnm{Hussein}, \binits{S.A.}},
\bauthor{\bsnm{Amr}, \binits{B.}},
\bauthor{\bsnm{Refaat}, \binits{S.}},
\bauthor{\bsnm{Shorim}, \binits{N.}},
\bauthor{\bsnm{Ghanim}, \binits{T.M.}}:
\bctitle{Hieroglyphs language translator using deep learning techniques (scriba)}.
In: \bbtitle{2022 2nd International Mobile, Intelligent, and Ubiquitous Computing Conference (MIUCC)}
(\byear{2022}).
\doiurl{10.1109/miucc55081.2022.9781784}
\end{bchapter}
\endbibitem

\bibitem{Sobhy2023}
\begin{botherref}
\oauthor{\bsnm{Sobhy}, \binits{A.}},
\oauthor{\bsnm{Helmy}, \binits{M.}},
\oauthor{\bsnm{Khalil}, \binits{M.}},
\oauthor{\bsnm{Elmasry}, \binits{S.}},
\oauthor{\bsnm{Boules}, \binits{Y.}},
\oauthor{\bsnm{Negied}, \binits{N.}}:
An ai based automatic translator for ancient hieroglyphic language-from scanned images to english text.
IEEE Access
(2023)
\end{botherref}
\endbibitem

\bibitem{Tang2016}
\begin{bchapter}
\bauthor{\bsnm{Tang}, \binits{Y.}},
\bauthor{\bsnm{Peng}, \binits{L.}},
\bauthor{\bsnm{Xu}, \binits{Q.}},
\bauthor{\bsnm{Wang}, \binits{Y.}},
\bauthor{\bsnm{Furuhata}, \binits{A.}}:
\bctitle{Cnn based transfer learning for historical chinese character recognition}.
In: \bbtitle{12th IAPR Workshop on Document Analysis Systems (DAS)},
pp. \bfpage{25}--\blpage{29}.
\bpublisher{IEEE}, \blocation{???}
(\byear{2016}).
\doiurl{10.1109/das.2016.52}
\end{bchapter}
\endbibitem

\bibitem{AlNoori2020}
\begin{bchapter}
\bauthor{\bsnm{Al-Noori}, \binits{A.H.}},
\bauthor{\bsnm{Talib}, \binits{M.}},
\bauthor{\bsnm{Harbi}, \binits{J.}}:
\bctitle{The classification of ancient sumerian characters using convolutional neural network}.
(\byear{2020}).
\bcomment{103770.pdf}.
\burl{https://www.scitepress.org}
\end{bchapter}
\endbibitem

\bibitem{AsmaaSobhy2023}
\begin{bchapter}
\bauthor{\bsnm{Sobhy}, \binits{A.}},
\bauthor{\bsnm{Helmy}, \binits{M.}},
\bauthor{\bsnm{Khalil}, \binits{M.}},
\bauthor{\bsnm{Elmasry}, \binits{S.}},
\bauthor{\bsnm{Boules}, \binits{Y.}},
\bauthor{\bsnm{Negied}, \binits{N.}}:
\bctitle{An ai based automatic translator for ancient hieroglyphic language—from scanned images to english text},
pp. \bfpage{38796}--\blpage{38804}.
\bpublisher{IEEE}, \blocation{???}
(\byear{2023}).
\doiurl{10.1109/ACCESS.2023.3267981}
\end{bchapter}
\endbibitem

\bibitem{AdobeStock}
\begin{botherref}
Hieroglyphics.
\url{https://stock.adobe.com/eg/search?k=hieroglyphics}
\end{botherref}
\endbibitem

\bibitem{AlamyStock}
\begin{botherref}
Egyptian Hieroglyphs.
\url{https://www.alamy.com/stock-photo/egyptian-hieroglyphs.html?sortBy=relevant}
\end{botherref}
\endbibitem

\bibitem{Piankoff1969}
\begin{bbook}
\bauthor{\bsnm{Piankoff}, \binits{A.}}:
\bbtitle{The Pyramid UNAs}.
\bsertitle{Bollingen Series}.
\bpublisher{Princeton Univ. Press},
\blocation{Princeton, NJ, USA}
(\byear{1969})
\end{bbook}
\endbibitem

\bibitem{RoseGitHub}
\begin{botherref}
\oauthor{\bsnm{Rose}, \binits{F.}}:
Egyptian Translation
(2020).
\url{https://github.com/fayrose/EgyptianTranslation/tree/main}
\end{botherref}
\endbibitem

\bibitem{Hough2015}
\begin{botherref}
\oauthor{\bsnm{Hassanein}, \binits{A.S.}},
\oauthor{\bsnm{Mohammad}, \binits{S.}},
\oauthor{\bsnm{Sameer}, \binits{M.}},
\oauthor{\bsnm{Ragab}, \binits{M.E.}}:
A survey on hough transform, theory, techniques and applications.
ArXiv
\textbf{abs/1502.02160}
(2015)
\end{botherref}
\endbibitem

\bibitem{Gong2018}
\begin{barticle}
\bauthor{\bsnm{Gong}, \binits{X.}},
\bauthor{\bsnm{Su}, \binits{H.}},
\bauthor{\bsnm{Xu}, \binits{D.}}, \betal:
\batitle{An overview of contour detection approaches}.
\bjtitle{Int. J. Autom. Comput.}
\bvolume{15},
\bfpage{656}--\blpage{672}
(\byear{2018}).
\doiurl{10.1007/s11633-018-1117-z}
\end{barticle}
\endbibitem

\bibitem{he2016deep}
\begin{bchapter}
\bauthor{\bsnm{He}, \binits{K.}},
\bauthor{\bsnm{Zhang}, \binits{X.}},
\bauthor{\bsnm{Ren}, \binits{S.}},
\bauthor{\bsnm{Sun}, \binits{J.}}:
\bctitle{Deep residual learning for image recognition}.
In: \bbtitle{Proceedings of the IEEE Conference on Computer Vision and Pattern Recognition},
pp. \bfpage{770}--\blpage{778}
(\byear{2016})
\end{bchapter}
\endbibitem

\bibitem{FSMNLP2011}
\begin{botherref}
Intersection of multitape transducers vs. cascade of binary transducers: The example of egyptian hieroglyphs transliteration.
In: Proceedings of the 9th International Workshop on Finite State Methods and Natural Language Processing,
pp. 74--82.
Association for Computational Linguistics,
Blois, France
(2011).
c©2011 Association for Computational Linguistics
\end{botherref}
\endbibitem

\bibitem{OurDataset}
\begin{botherref}
Dataset.
\url{https://github.com/mariamsk10/HieroGlyphTranslator.git}
\end{botherref}
\endbibitem

\end{thebibliography}

\end{document}